\documentclass[letterpaper, 10 pt, conference]{ieeeconf}  

\IEEEoverridecommandlockouts                              

\usepackage{graphics} 
\usepackage{amsmath} 
\usepackage{amssymb}  
\usepackage[capitalise]{cleveref}
\usepackage{graphicx}
\usepackage{xcolor}
\usepackage{subcaption}
\usepackage{amssymb}
\usepackage{pifont}

\title{\LARGE \bf
Conditional Flow-VAE for Safety-Critical Traffic Scenario Generation
}

\author{Zimu Gong$^{3,\dagger,\star}$, Brian Zhaoning Zhang$^{4,\dagger,\star}$, Chris Zhang$^{1,2,\star}$, Kelvin Wong$^{1,2}$, and Raquel Urtasun$^{1,2}$%
\thanks{The authors are with $^{1}$Waabi Innovation Inc, $^{2}$University of Toronto, $^{3}$University of Michigan-Ann Arbor, $^{4}$University of Waterloo.}%
\thanks{$^{\dagger}$Work done during affiliation with Waabi.}%
\thanks{$^{\star}$Equal contribution.}%
}

\begin{document}

\maketitle
\thispagestyle{empty}
\pagestyle{empty}


\begin{abstract}

    Safety-critical scenarios are essential for the development of 
    autonomous vehicles (AVs) but are rare in real-world driving data. While simulation offers a way to generate
    such scenarios, manually designed test cases lack scalability, and
    adversarial optimization often produces unrealistic behaviors. In this work,
    we introduce a conditional latent flow matching approach for scalable and
    realistic safety-critical scenario generation. Our method uses distribution matching to transform nominal scenes into
    safety-critical rollouts. Furthermore, we demonstrate that incorporating
    both simulation and real-world data enables our framework to efficiently
    generate diverse, data-driven scenarios. Experimental results highlight that
    our approach is able to more consistently and realistically generate
    novel safety-critical scenarios,
    making it a valuable tool for training and benchmarking AV
    systems.

\end{abstract}
\section{INTRODUCTION}

Safety-critical scenarios play a central role in the development of autonomous
vehicles (AVs). Rare events such as sudden cut-ins, near-miss
interactions, or unexpected braking are precisely the situations where an AV’s
decision-making and planning policies are most challenged. Robust performance is essential, yet exposing AV systems to
these conditions in the real world is costly and dangerous. Simulation is therefore critical: it enables evaluation under
safety-critical conditions before deployment, reducing risk and accelerating
development.

However, acquiring a sufficiently diverse and realistic set of safety-critical
scenarios for simulation remains a major challenge.
Traditionally, simulation-based approaches typically rely on heavy human curation.
For example, \cite{kusano2022collisionavoidancetestingwaymo} identifies potential hazardous events in
their ODD and then recreates them in simulation, and
\cite{scanlon2021waymoreconstruction} manually reconstructs safety-critical
scenarios from police crash reports.
However, this approach is far too tedious to scale efficiently
and cost-effectively.
Automated methods like adversarial optimization
where agents are constructed to deliberately collide with the ego vehicle
can be more easily scaled. However, real traffic participants
are not inherently adversarial,  and such methods neglect
the prevalence of near-miss situations that pose genuine challenges for autonomy
systems.
In both cases, the resulting scenarios may not reflect the
statistics of real world driving.

Alternatively, \emph{distribution matching} approaches are a promising
avenue towards traffic simulation that matches the real world.
However, distribution matching approaches like imitation learning
are notoriously data hungry, but real safety critical scenarios
are inherently rare and difficult to obtain.
Standard traffic simulation models are trained predominantly on
nominal data and thus are naturally biased toward reproducing nominal behaviors rather
than generating safety-critical ones.
Unfortunately, overly upsampling the limited number of safety-critical scenarios
can easily lead to overfitting.

\begin{figure}
    \includegraphics[width=\linewidth]{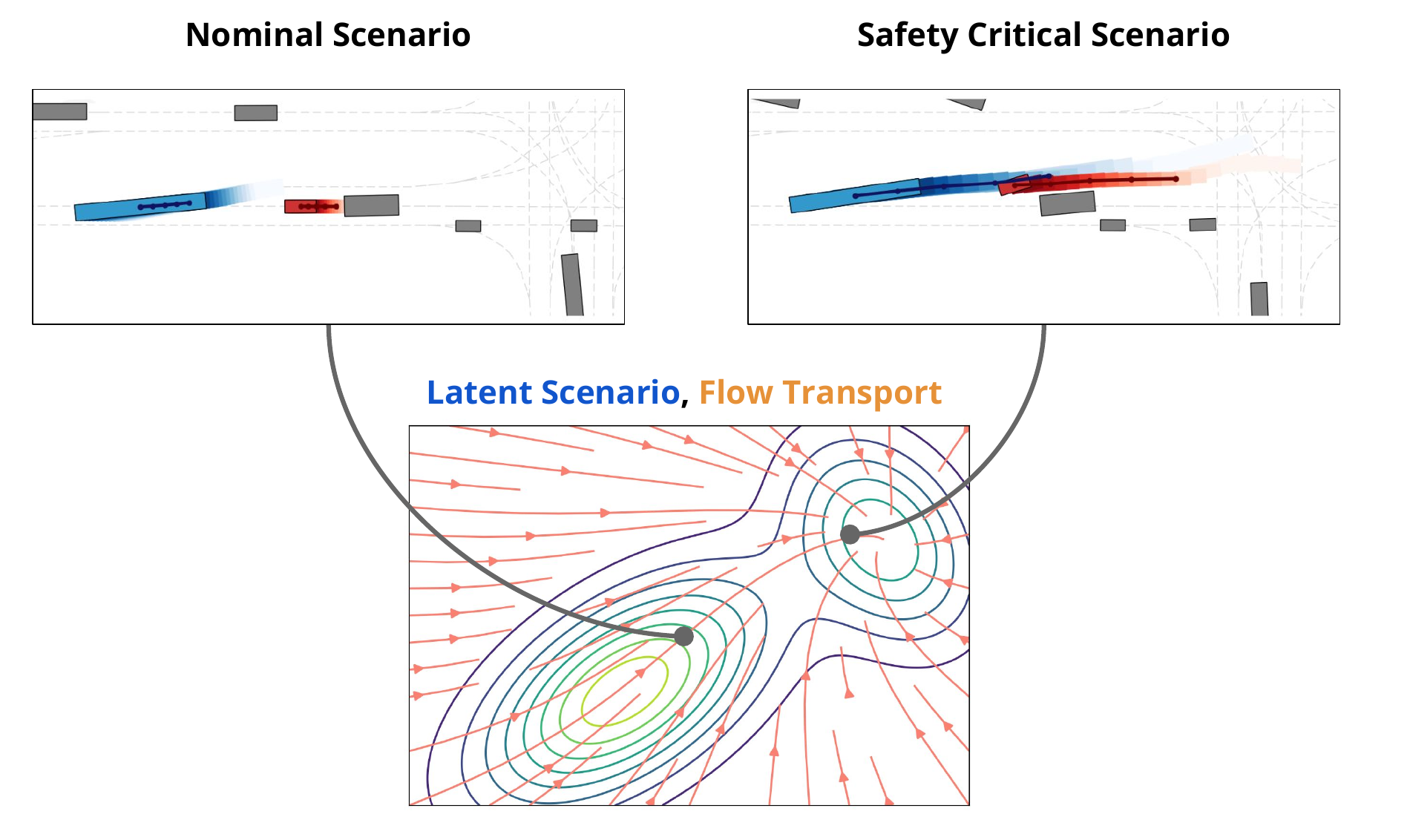}
    \caption{Our method learns a latent space over scenarios and flow transport to map the
        nominal distribution to the safety-critical distribution.}
    \label{fig:teaser}
\end{figure}
Towards a data-driven approach to generating safety-critical scenarios
while preserving behavioral realism, we propose our conditional flow VAE.
We supplement the limited number of high-fidelity real safety critical scenarios
with lower fidelity but easily scalable synthetic data,
and design a model which can maximally take advantage of this data mixture.
Specifically, our model captures the semantics of a variety of driving
scenarios with a conditional VAE encoder and uses a flow matching transformer
to transform the VAE latents from the nominal distribution to the safety-critical
counterpart.
Then we utilize the VAE decoder to produce safety-critical rollouts
from nominal driving scenarios, bridging the gap between rare real-world
events and scalable synthetic generation.
By incorporating both real-world and
simulated data, our approach captures the realism of human driving while
supplementing the long tail with synthetic diversity.
Empirically, our flow approach outperforms alternative conditioning baselines.
Furthermore, we enable controllable scenario difficulty by conditioning on automatically generated
heuristic labels, allowing systematic evaluation across varying levels of
criticality.

To summarize, our contributions are threefold:

\begin{itemize}
    \item A generative framework for safety-critical scenario
          generation based on conditional latent flow matching.

    \item Realism through distribution alignment, enabling transitions from
          nominal to safety-critical outcomes without adversarial artifacts.

    \item Data efficiency and controllability, achieved through the integration
          of real and simulated data along with difficulty-conditioned generation.
\end{itemize}

Together, these advances provide a practical and principled way to generate
realistic safety-critical scenarios.

\section{RELATED WORK}

\subsection{Challenging Scenario Generation}

Scenario generation includes a two-fold objective of actor placement and actor
roll-out. The former objective generates a new scenario initialization from
scratch or modifies existing scenarios. It often specifies the actor placement
and initialization states so that the downstream simulation may yield
safety-critical outcomes \cite{feng2023trafficgenlearninggeneratediverse},
\cite{tan2023languageconditionedtrafficgeneration}. In this paper, we focus on
generating safety-critical scenarios from a nominal initialization with a
roll-out model, where we train a model to directly control the actor maneuvers
to be applied upon any nominal initialization. 

Early approaches rely on manually designed scenarios and heuristic rules
\cite{PhysRevE.62.1805}, often embedded in simulation platforms such as CARLA
\cite{DBLP:journals/corr/abs-1711-03938}. These methods offer clear control over
vehicle maneuvers by manually specifying the planning trajectory and kinematic
constraints. However, they are often limited in scalability and diversity, as
each scenario must be explicitly scripted by engineers. The scenario parameters
also need to be decided carefully to ensure that there is no collision or other
undesired behaviors. 

The advent of data-driven approaches allows the machine learning model design
that focuses on learning from mass-scale driving datasets like WOMD
\cite{Kan_2024_icra}. With generative machine learning model architectures like
VAE, autoregressive models and diffusion models \cite{10089194}, these methods
are often trained on behavior cloning objectives that allow high reconstruction
L2 scores on the eval data \cite{wu2024smartscalablemultiagentrealtime},
\cite{zhong2022guidedconditionaldiffusioncontrollable},
\cite{xu2022bitsbilevelimitationtraffic}. They also utilize common knowledge
like collision loss and traffic signals to enhance 
realism~\cite{suo2021trafficsimlearningsimulaterealistic,zhang2023learning}. 
However, the behavior cloning objective implies that the model is fitted to the distribution
of the training dataset, where most of the data are
nominal~\cite{zhai2023rethinkingopenloopevaluationendtoend}, which hinders the
ability to diversify towards safety-critical rollouts. 

Another approach is to apply an adversarial objective to the actors in the
scenario. Many of those apply similar strategies where they manipulate the
learned representation of the actors in the latent space. 
STRIVE~\cite{rempe2022strive} utilizes an optimization objective on the latent space of the CVAE.
Other methods use reinforcement learning based editing
\cite{liu2024safetycriticalscenariogenerationreinforcement},
\cite{zhang2024learningdriveasymmetricselfplay},
\cite{wang2023advsimgeneratingsafetycriticalscenarios}. The actors in the
scenario are often set with an adversarial objective to cause collisions with
the ego actor. While effective at exposing the weakness of planners, these
methods often compromise realism: real traffic actors are not inherently
adversarial, thus the resulting trajectories may be unnatural. Besides, the
method also requires the presence of a planner module for the ego vehicle,
leading to more complex training architecture. 

\subsection{VAE and Flow Matching Model}

Variational Autoencoder models are useful in representation learning, where it
efficiently compresses the input features into a latent space, so that a decoder
can utilize them for downstream tasks
\cite{kingma2022autoencodingvariationalbayes}. It is desired to manipulate in
the latent space to achieve specific objectives. Research in computer vision 
has shown that the representation space can be decomposed into subcomponents
and then used for domain translation
\cite{huang2018multimodalunsupervisedimagetoimagetranslation}. In the context of
autonomous driving, VAE encodes traffic scenarios efficiently conditioned on the
past states and high-level scenario information, and a decoder head is usually
applied to generate per-actor roll-outs. Previous work
\cite{bairouk2024exploringlatentpathwaysenhancing} have shown that the VAE
embedding space contains useful information that is valuable for interpretable
maneuver generation. In our work, we show that the nominal and safety-critical
scenarios reside in different subsections of the latent space and a transfer map
can be learned to map the nominal latents to the safety-critical latents. 

A flow matching model is a type of generative model that learns to transform
samples from a simple base distribution into samples from a
complex target distribution following a continuous flow
\cite{lipman2023flowmatchinggenerativemodeling}. Instead of learning a discrete
sequence of transformations (like normalizing flows) or stepping with random
noises (like diffusion), flow matching learns a vector field that describes
explicitly how data should move between source and target distributions over
time. During inference, the flow model uses a sampler to step through the
timesteps from 0 to 1 along the learned field to map the sample from the source
distribution to the target distribution. Key developments, including Rectified
Flow \cite{liu2022flowstraightfastlearning}, allows flowing from an arbitrary
source distribution instead of the standard Gaussian. The method is widely
applied to various tasks like text-to-image generation
\cite{esser2024scalingrectifiedflowtransformers}, and robotics
\cite{black2024pi0visionlanguageactionflowmodel}. It is also easy to apply
conditioning on the flow model to boost the performance
\cite{nichol2022glidephotorealisticimagegeneration}. Compared to diffusion, the
computational efficiency and stable training objective make it preferable
for learning the transfer from nominal scenarios into safety-critical ones
within the latent space. 

\section{PRELIMINARIES}

\begin{figure*}[t]
    \centering
    \includegraphics[width=\textwidth]{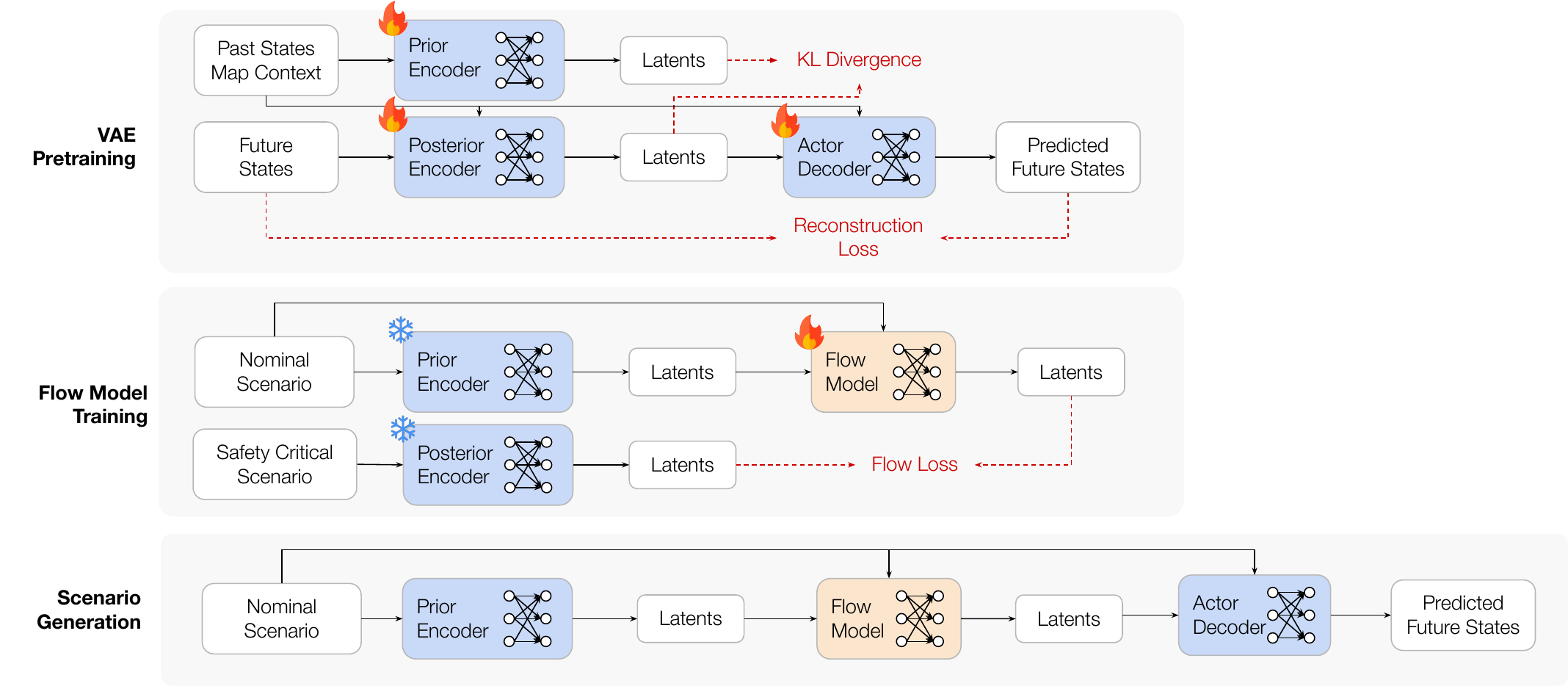}
    \caption{
        \textbf{Conditional Flow VAE}. 
        We first pretrain the VAE model on a mixture of data. 
        Next, the VAE model is frozen and a flow model learns 
        to map nominal latents to safety-critical latents.
        During inference, the flow transformer takes the prior latent and predicts the posterior latent.
        All the latents are passed through the CVAE decoder to generate the final actor states. }
    \label{figurelabel}
\end{figure*}

\subsection{Traffic Modeling}
We define the problem scope as generating a traffic scenario with $N$ actors in
a finite horizon of $T$ time steps. We use $Y^t = \{y_1^t, y_2^t,...,y_N^t\}$ to
denote the actor states at time $t$. For each vehicle state we define $y_i^t =
    (b_x, b_y, b_z, b_\theta, b_v, b_l, b_w, b_h)$, which describes the 3D position,
yaw, velocity, and the length, width, and height of a vehicle's bounding box.
The model observes the high-definition map $\mathbf{M}$ and past $H$ states
$Y^{-H:0}$ and outputs the vehicle control action sequence $A^t=\{a_1^t,
    a_2^t,...,a_n^t\}$, where $a_i^t = (a_{accel}, a_{steer})$. The goal of traffic
modeling is typically to learn to model the distribution over future actor
states.
\begin{equation}
    p(Y^{1:T} | \mathbf{M}, Y^{-H:0})
\end{equation}

\subsection{Rectified Flow}
Given observations of two distributions
$p_0$, $p_1$ on $\mathbb{R}^d$ we wish to find a transport map
$T: \mathbb{R}^d \rightarrow \mathbb{R}^d$
such that $X_1:=T(X_0) \sim p_1$ when $X_0\sim p_0$.
A rectified flow is an ODE on time $t\in[0, 1]$
\begin{equation}
    dZ_t = v_\theta(Z_t, t)dt
\end{equation}
where $v_\theta:\mathbb{R}^d \rightarrow \mathbb{R}^d$ is a velocity field that
learns to drive the flow along the linear path from $X_0$ to $X_1$ by optimizing
\begin{equation}
    \int_0^1 \mathbb{E} \left[|| (X_1 - X_0) - v(X_t,t)||^2 \right]dt
\end{equation}
where $X_t = tX_1 + (1-t)X_0$. After learning, $v_\theta$ can be used to sample
from $p_1$ given samples from $p_0$.

\subsection{Variational Autoencoder}
The variational autoencoder is a latent variable approach to generative modeling
\begin{equation}
    p(x) = \int_z p(x|z)p(z)dz
\end{equation}
where $z$ is some latent variable meant to capture unobserved aspects of the
generative process. An encoder $q_\theta(z|x)$, decoder $p_\theta(x|z)$ and
prior $p_\theta(z)$ can be jointly learned to optimize the evidence lower bound:
\begin{equation}
    \mathbb{E}_{z\sim q_\theta}\left[ \log p_\theta (x| z)\right] + \mathcal{D}_{\text{KL}}\left(q_\theta(\cdot|x) \;||\; p_\theta(\cdot)\right)
\end{equation}
where $\mathcal{D}_{\text{KL}}$ is the Kullback-Leibler divergence. After learning,
sampling from $p(x)$ amounts to sampling from the prior followed by the
decoder.
VAEs can be extended to support conditional generation~\cite{sohn2015learning}
by extending the encoder, decoder and prior to be conditional distributions as well,
e.g. $q_\theta(z | x, c)$, $p_\theta(x |z, c)$, $p_\theta(z | c)$ respectively.

\section{METHODOLOGY}
The overall framework of the conditional flow VAE is depicted in Fig.
\ref{figurelabel}.

\subsection{Latent flow matching}
We take a distribution matching approach for safety-critical scenario
generation. Let $p_N$ be the nominal distribution of traffic scenarios for which
we have many samples, and $p_S$ be the safety-critical distribution, for which
we have comparatively fewer samples. Our approach learns a flow from $p_N$ to
$p_S$. Doing so allows us to learn to sample new scenarios from $p_S$ by
transporting samples from $p_N$.

To begin, we follow \cite{suo2021trafficsimlearningsimulaterealistic} and
model general traffic scenarios using a conditional VAE:
\begin{multline}
    p(Y^{1:T} | \mathbf{M}, Y^{-H:0}) = \\
    \int_Z  p_\theta(Y^{1:T} | \mathbf{M}, Y^{-H:0}, Z) p_\theta (Z | \mathbf{M}, Y^{-H:0}).
\end{multline}
In this case, the encoder (posterior) is given as $q_\theta (Z |  \mathbf{M},
    Y^{-H:0}, Y^{1:T})$. Note that we learn the conditional VAE on the
\textit{mixture} of $p_N$ and $p_S$, meaning the model is trained on samples
from both distributions.

We now define the \emph{safety-critical} aggregate posterior as
\begin{multline}
    q^{S}_\theta(Z|\mathbf{M}, Y^{-H:0}) = \\
    \mathbb{E}_{p_S(Y^{1:T} |  \mathbf{M}, Y^{-H:0})} \left[ q_\theta (Z |  \mathbf{M}, Y^{-H:0}, Y^{1:T}) \right].
\end{multline}
Intuitively, this is the distribution over $Z$ for safety-critical scenarios.
We sample from $q^{S}$ using the VAE posterior on safety-critical scenarios.
The flow model then aims to learn a transport between the prior and this
safety-critical latent distribution by optimizing
\begin{equation}
    \int_0^1 \mathbb{E} \left[|| (Z_1 - Z_0) - v(Z_t,t)||^2 \right]dt
\end{equation}
where $Z_t = tZ_1 + (1-t)Z_0$, and $Z_1\sim q_\theta^{S}$ and
$Z_0 \sim p_\theta (Z | \mathbf{M}, Y^{-H:0})$.
This objective trains the model to map prior latents
to safety-critical latents.

There are several advantages to our approach. By training the VAE on the mixture
of nominal and safety-critical data, we are able to learn better overall
realistic driving by making use of all data, as opposed to learning only on
safety-critical data. However, the explicit flow objective allows us to steer
our sampling towards the safety-critical distribution. Compared to trajectory
space, flowing in latent space also allows us to control the degree of safety
criticality by doing a partial flow (e.g. until $t=0.5$), since the decoder
still maps intermediate results in latent space to plausible futures.




\subsection{Training Recipe}
We follow a two-stage training procedure inspired by latent space manipulation
practices in computer
vision~\cite{razavi2019generatingdiversehighfidelityimages}. In the first stage,
we train a conditional VAE on both nominal and safety-critical scenarios to
establish a stable and semantically meaningful latent representation for both
types. In the second stage, we train the flow model exclusively on
safety-critical scenarios to learn the distributional transport from nominal to
safety-critical latents. We found that this staged design is essential for reliable
convergence. This is because empirically we observed that for a
single stage end-to-end training approach, the prior and posterior distributions
of the VAE undergo large shifts in the early iterations, while the flow model
simultaneously attempts to learn the mapping with high learning rates. As a
result, the flow model is effectively trained on a non-stationary target, which
often leads to instability~\cite{he2019lagginginferencenetworksposterior}. By
decoupling the stages, the VAE first provides a fixed latent space learned under
a combination of imitation and traffic-compliance objectives. The subsequent
flow model is then trained with a flow objective, which benefits from theoretical
convergence guarantees under a fixed latent
space~\cite{jiao2024convergenceanalysisflowmatching}.

\subsection{Architecture}
We now describe the neural network architecture used
for the different components of our approach.

\paragraph{Backbone}\label{sec:backbone}
A transformer~\cite{vaswani2017attention} based architecture is used as the backbone
network for the VAE and flow transformer.
We adopt several common techniques used in traffic modelling.
We first extract state features from each actor using a simple MLP.
Map features are extracted using an off-the-shelf map encoder~\cite{cui2022gorela}.
Map and actor state features are then augmented with
PairPose relative positional features~\cite{cui2022gorela} allowing for viewpoint-invariance.
Our transformer comprises interleaved actor-to-map, actor-to-actor and actor-to-time attention layers~\cite{ngiam2021scene};
relative positional encodings~\cite{cui2022gorela,zhong2022guidedconditionaldiffusioncontrollable,zhou2022hivt}
between actors are used to preserve viewpoint-invariance.
To save on computational cost, actor-to-actor and actor-map attention is limited
to the top-$k$ closest actors or lane graph nodes, essentially forming a local context
for each actor.

\paragraph{VAE}
Following prior work on multi-agent traffic
simulation~\cite{suo2021trafficsimlearningsimulaterealistic,rempe2022strive}, we employ a
conditional variational autoencoder (CVAE) approach to learn latent embeddings that
capture rich scene semantics and the multi-agent interactions.
The prior $p_\theta (Z | \mathbf{M}, Y^{-H:0})$
and posterior $q_\theta (Z | \mathbf{M}, Y^{-H:0}, Y^{1:T})$
use the same backbone described above, differing only in
the number of timesteps of actor states observed, sharing the map encoder.
Note that, similar to \cite{suo2021trafficsimlearningsimulaterealistic}, we predict
a separate latent for each actor. The decoder also uses the same backbone;
the latent is fused into the actor feature, and a steering and acceleration is predicted
per actor.

\begin{table*}[t]
    \caption{\textbf{Realistic Safety-critical Scenario Generation}.
        We evaluate against baselines on our held out set
        of real safety critical scenarios, and
        obtain the highest near miss \% (valid safety critical scenario),
        while having high distributional similarity.
    }
    \label{table:benchmark}
    \begin{center}
        \begin{tabular}{c | c c c c | c c c}
            \hline
                           &                      &                           &                       &                          & \multicolumn{3}{c}{Distribution JSD}                                          \\
                           & minSTTC $\downarrow$ & Near Miss (\%) $\uparrow$ & SCR (\%) $\downarrow$ & Displ Error $\downarrow$ & Velocity $\downarrow$                & Accel $\downarrow$ & Jerk $\downarrow$ \\
            \hline
            VAE            & 2.927                & 22.9                      & 0.7                   & 5.71                     & 0.245                                & 0.101              & 0.017             \\
            VAE + Curation & 3.776                & 31.2                      & 0.8                   & 7.32                     & 0.278                                & 0.136              & 0.019             \\
            STRIVE         & 1.885                & 30.2                      & 7.0                   & 7.22                     & 0.277                                & 0.151              & 0.013             \\
            FlowVAE        & 2.190                & 45.8                      & 1.6                   & 6.51                     & 0.262                                & 0.131              & 0.021             \\
            \hline
        \end{tabular}
    \end{center}
\end{table*}
\begin{figure*}[t]
    \centering
    \begin{subfigure}[t]{0.33\textwidth}
        \centering
        \includegraphics[width=\textwidth,trim={0 2cm 30cm 2cm},clip]{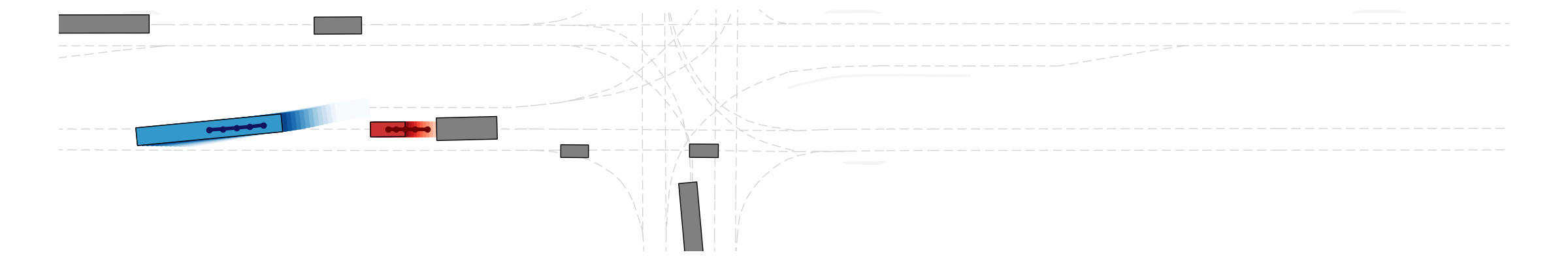}
        \includegraphics[width=\textwidth,trim={0 1cm 30cm 2cm},clip]{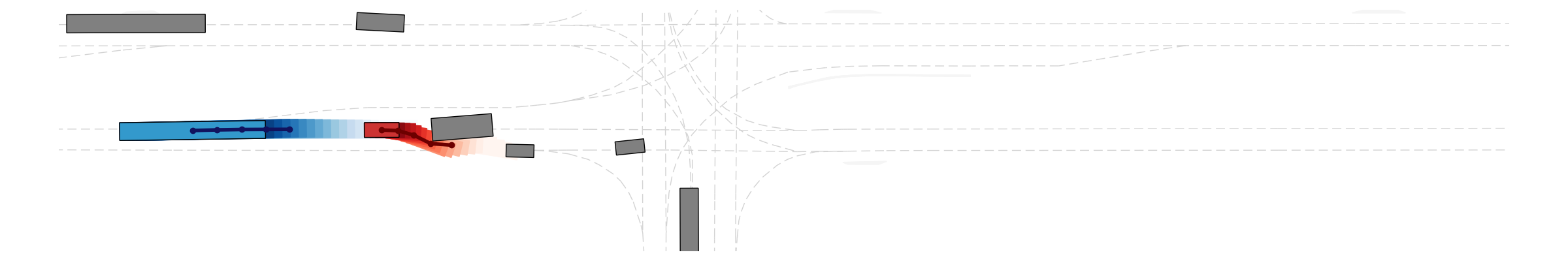}
        \includegraphics[width=\textwidth,trim={0 1cm 30cm 2cm},clip]{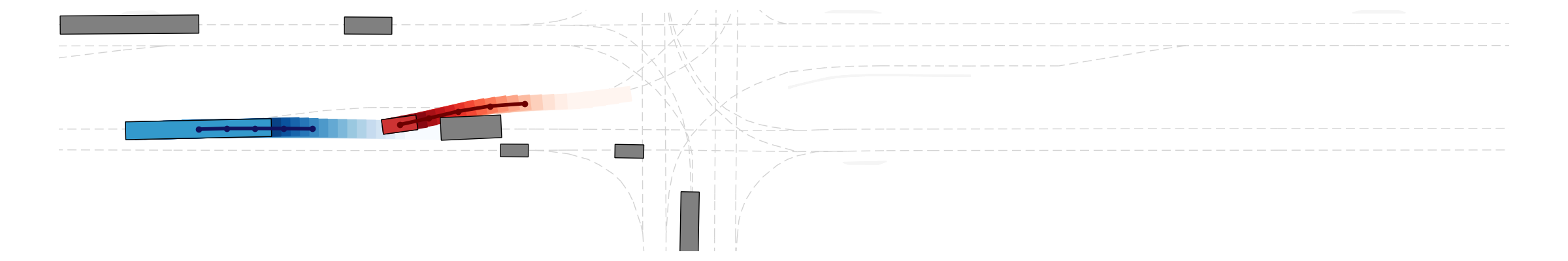}
        \includegraphics[width=\textwidth,trim={0 1cm 30cm 2cm},clip]{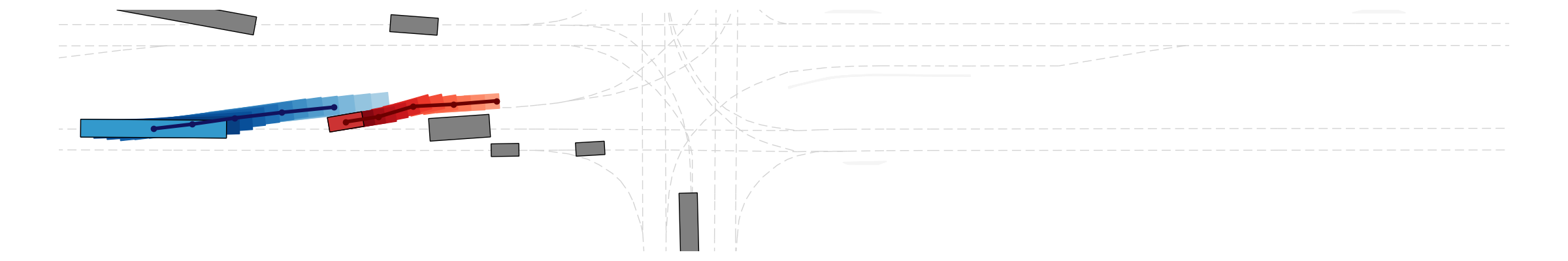}
        \caption{Urban cut-in scenario}
    \end{subfigure}%
    ~
    \begin{subfigure}[t]{0.33\textwidth}
        \centering
        \includegraphics[width=\textwidth,trim={0 1cm 30cm 2cm},clip]{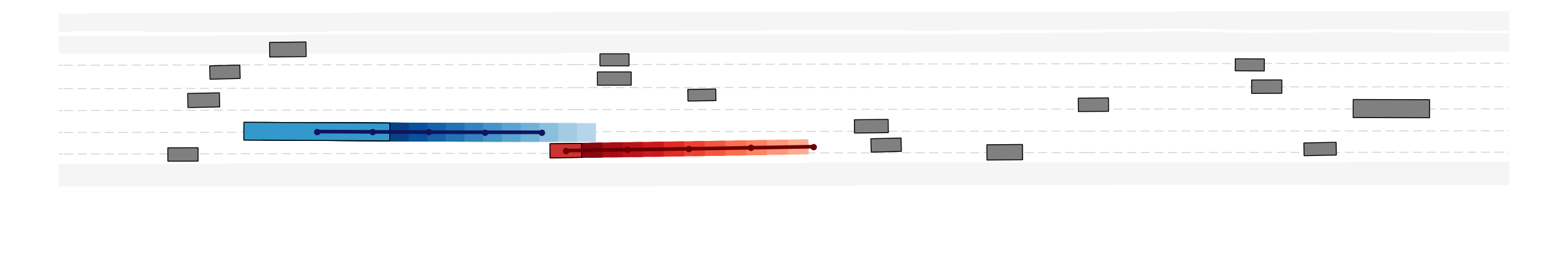}
        \includegraphics[width=\textwidth,trim={0 1cm 30cm 2cm},clip]{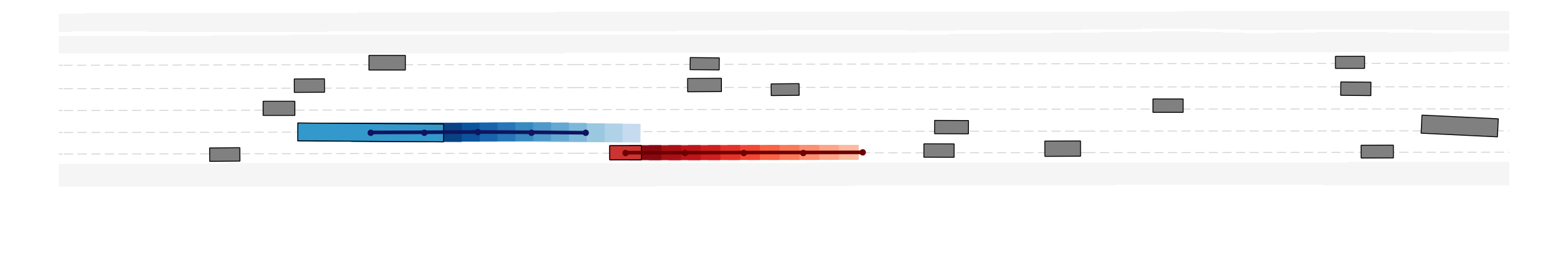}
        \includegraphics[width=\textwidth,trim={0 1cm 30cm 2cm},clip]{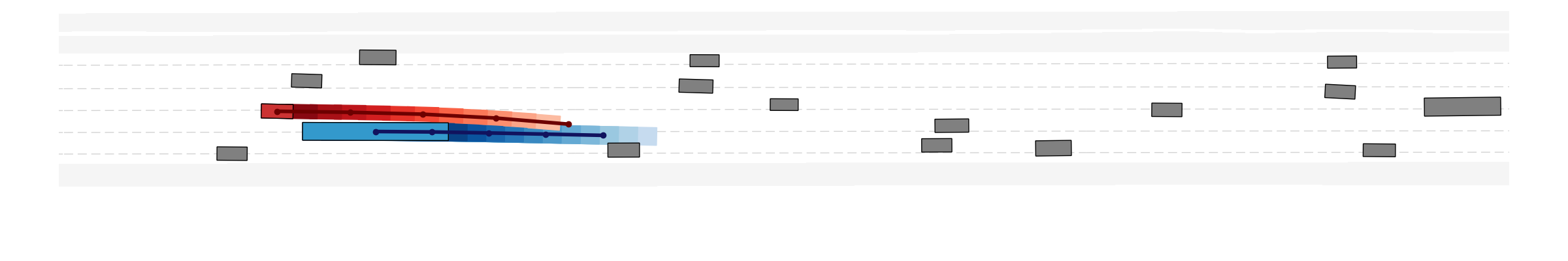}
        \includegraphics[width=\textwidth,trim={0 1cm 30cm 2cm},clip]{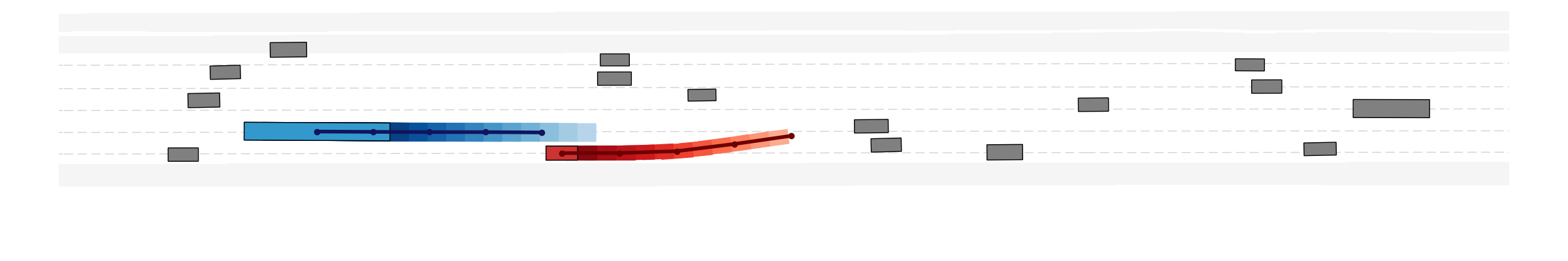}
        \caption{Highway cut-in scenario}
    \end{subfigure}%
    ~
    \begin{subfigure}[t]{0.33\textwidth}
        \centering
        \includegraphics[width=\textwidth,trim={22cm 1cm 7cm 2cm},clip]{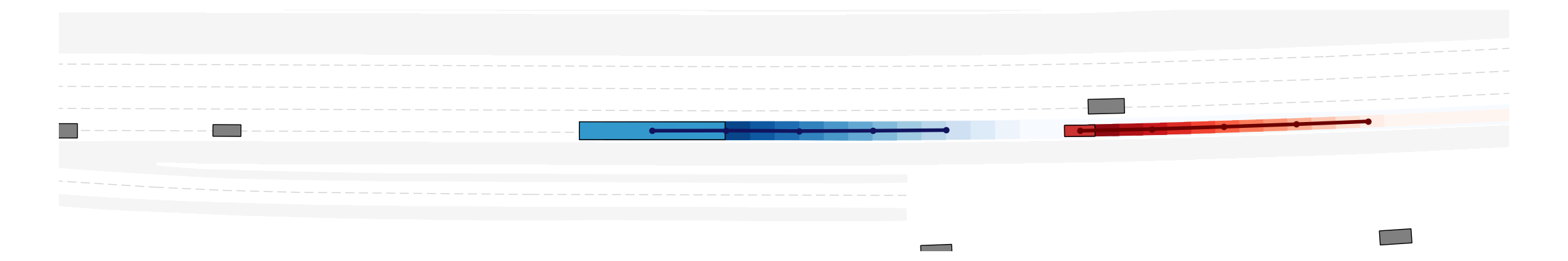}
        \includegraphics[width=\textwidth,trim={15cm 1cm 15cm 2cm},clip]{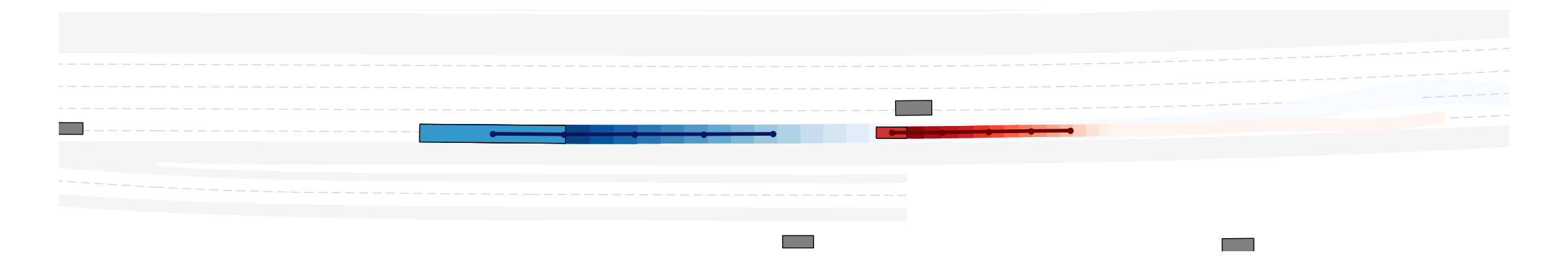}
        \includegraphics[width=\textwidth,trim={15cm 1cm 15cm 2cm},clip]{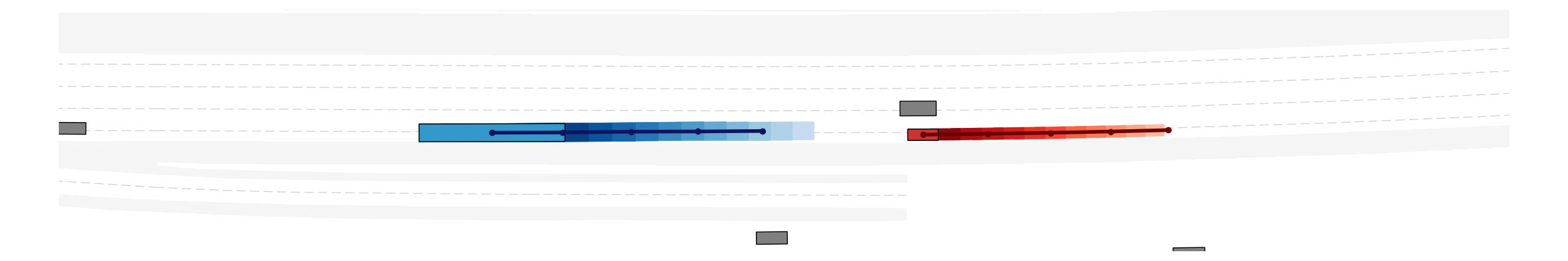}
        \includegraphics[width=\textwidth,trim={15cm 1cm 15cm 2cm},clip]{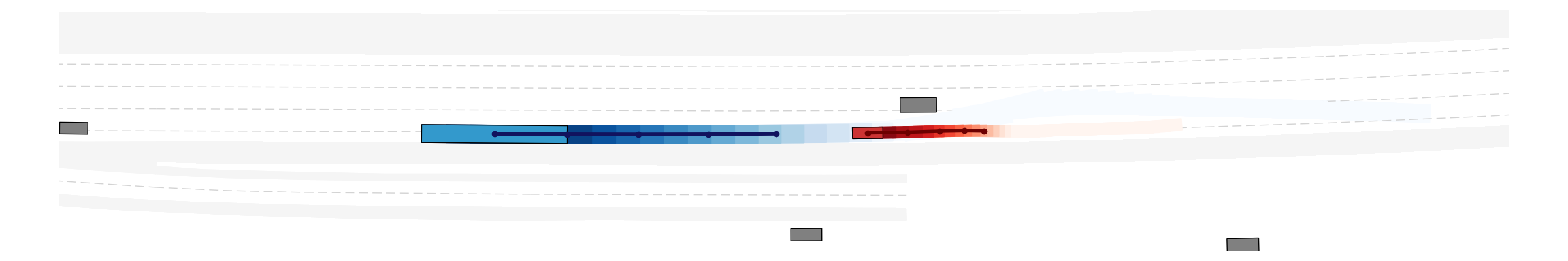}
        \caption{Highway hard-brake scenario}
    \end{subfigure}
    \caption{\textbf{Qualitatives}. From top to bottom: original nominal scenario, VAE reconstruction, STRIVE, our model.}
    \label{figure:demo_viz}
\end{figure*}
\paragraph{Flow model}
With the stable latent representation established from the CVAE, safety-critical
rollout generation requires transforming nominal latents into their
safety-critical counterparts. We frame this as a distribution matching problem:
safety-critical behaviors (e.g., hard braking, aggressive cut-ins) occupy
distinct subregions of the latent space, and our goal is to learn a mapping from
nominal priors to these critical submanifolds.

We use the same transformer backbone for our flow model.
Our transformer backbone is conditioned on the actor state feature
to implement flow matching in latent space.
We combine the flow matching
context along with the actor-level features to form the following input features:
\begin{equation}
    \begin{split}
        E_{actor} & = [X^t, E_T, E_x, E_c]                                                             \\
        E_T       & = \text{SinusoidalPE}(t_{denoise})                                                 \\
        E_x       & = t_{\text{denoise}} Z_{\text{posterior}} + (1-t_{\text{denoise}})Z_{\text{prior}} \\
        E_c       & = \text{MLP}(\text{ManeuverLabel}),
    \end{split}
\end{equation}
where $t_{\text{denoise}}$ is the denoising time step drawn from the uniform
distribution $U(0, 1)$. $X$ is actor state and map features.
$E_T$ is the sinusoidal positional encoding of the
denoising time step. $E_x$ is an interpolation between the prior latent
$Z_{\text{prior}}$ and the posterior latent $Z_{\text{posterior}}$.
Since the flow model only
conditions on the scene initialization, an optional maneuver indicator is
accepted to control the level of aggressiveness of maneuver. We classify each
scenario in the training dataset into one of the three categories: nominal,
safety-critical, and very safety-critical. The label is computed with
heuristics on vehicle kinematics and time-to-collision. $E_c$ is the maneuver
label projected with an MLP.


During inference, we discard the posterior encoder and use the prior encoder only
since ground-truth futures are unavailable. Given any initialization state, the
prior encoder produces a latent $Z_\text{prior}$, which is then transformed by the
Flow Transformer into a steered latent $Z_{pred}$. This latent is decoded
through the CVAE decoder to acquire the final actor states.

\subsection{Dataset}\label{sec:dataset}
Performing distribution matching between nominal and safety-critical rollouts
requires access to safety-critical scenarios observed in real driving logs.
However, due to their inherent sparsity in naturalistic datasets, none of the
widely used open-source autonomous driving corpora explicitly curate such
subsets. To address this limitation, we adopt a simulation–real data mixing
strategy that balances scalability with realism.
For real safety-critical scenarios, we conduct targeted data mining over a catalogue of real driving logs.
We extract approximately 500 unique scenarios with challenging situations;
e.g., actors perform abrupt cut-ins, aggressive braking, etc.
The result of this data mining is a set of smaller scale but high-fidelity
demonstrations of realistic human driving behaviors under challenging conditions.


To supplement the real safety critical scenarios, we leverage
simulation.
Our goal is to generate realistic scenarios
which introduce diverse behaviors that can help transfer
to real safety critical scenarios.
We use existing deep-learning based traffic simulation models for nominal traffic,
with the addition of
a ``hero actor'' selected among existing actors with heuristics,
or additionally
inserted into the scene.
It is parameterized by Intelligent Driver Model (IDM)
heuristics and programmed to execute either a cut-in or a hard braking maneuver.
Overall, this procedure provides reasonably fine-grained control
over desired maneuvers and generates many new scenarios.
We leverage rejection sampling to throw away simulations that fail a small
set of basic checks due to a failure in the heuristics.
This allows us to generate an order of magnitude more safety critical
scenarios than we have mined from real logs.
We also found that the diversity of the simulation generated scenarios helps
improve the model performance. We created three versions of simulation data with
different heuristics: one with the most safety-critical maneuvers, one with
kinematic constraints (deceleration, TTC, etc.) approximately tuned to the real
data.
It turns out that using both versions of
simulation data achieves the best performance.
While the resulting behaviors themselves remain limited in diversity,
empirically we will show that they still provide benefits
and partially transfer to the real evaluation set.

During training, we blend real and simulated scenarios using a hyperparameter
$\alpha_{real}$, which determines the relative proportion of real and sim data.
Each training batch has an $\alpha_{real}\%$ chance to draw a sample from the
real dataset, and a $(1-\alpha_{real})\%$ chance to draw from a sim sample. This
mechanism allows us to smoothly adjust the balance between realism and
scalability, and to study the effect of sim--real composition on downstream
performance.
As real safety-critical data
alone is too scarce to provide sufficient coverage of the scenario space,
while sim-only data introduces a domain gap,
$\alpha_{real}$ serves as a way to balance the two, 
which we empirically validate in \cref{sec:ablate-data}.


\begin{table*}[t]
    \caption{
        \textbf{Conditional Flow Ablation.} We ablate the effect of
        our combined conditioning and flow approach. Our results show that
        both aspects are important, and together form the best results.
    }
    \label{table:model-ablate}
    \begin{center}
        \begin{tabular}{c c | c c c c | c c c}
            \hline
                       &              &                      &                           &                       &                          &                       & Distribution JSD   &                   \\
            Flow       & Conditioning & minSTTC $\downarrow$ & Near Miss (\%) $\uparrow$ & SCR (\%) $\downarrow$ & Displ Error $\downarrow$ & Velocity $\downarrow$ & Accel $\downarrow$ & Jerk $\downarrow$ \\
            \hline
                       &              & 2.927                & 22.9                      & 0.7                   & 5.71                     & 0.245                 & 0.101              & 0.017             \\
            \checkmark &              & 2.190                & 31.9                      & 1.7                   & 8.33                     & 0.279                 & 0.151              & 0.025             \\
                       & \checkmark   & 2.863                & 16.7                      & 1.7                   & 6.39                     & 0.268                 & 0.120              & 0.016             \\
            \checkmark & \checkmark   & 2.190                & 45.8                      & 1.6                   & 6.51                     & 0.262                 & 0.131              & 0.021             \\
            \hline
        \end{tabular}
    \end{center}
\end{table*}
\begin{table*}[t]
    \caption{\textbf{Simulation Data Transfers to Real.}
        We evaluate the effect of our simulation data and find that
        introducing a moderate amount of simulation data
        transfers to the real safety-critical distribution.
    }
    \label{table:real_mix}
    \begin{center}
        \begin{tabular}{c | c c c c | c c c}
            \hline
                       &                      &                           &                       &                          & \multicolumn{3}{c}{Distribution JSD}                                          \\
                       & minSTTC $\downarrow$ & Near Miss (\%) $\uparrow$ & SCR (\%) $\downarrow$ & Displ Error $\downarrow$ & Velocity $\downarrow$                & Accel $\downarrow$ & Jerk $\downarrow$ \\
            \hline
            Sim-only   & 3.052                & 39.5                      & 0.035                 & 7.98                     & 0.270                                & 0.166              & 0.038             \\
            10\% Real  & 2.944                & 39.6                      & 0.030                 & 7.97                     & 0.276                                & 0.151              & 0.029             \\
            20\% Real  & 2.357                & 45.8                      & 0.027                 & 7.49                     & 0.265                                & 0.139              & 0.028             \\
            40\% Real  & 1.945                & 58.3                      & 0.029                 & 7.19                     & 0.251                                & 0.146              & 0.032             \\
            60\% Real  & 2.568                & 58.3                      & 0.017                 & 7.31                     & 0.263                                & 0.136              & 0.028             \\
            100\% Real & 5.078                & 35.4                      & 0.024                 & 8.28                     & 0.284                                & 0.152              & 0.029             \\
            \hline
        \end{tabular}
    \end{center}
\end{table*}

\section{EXPERIMENTS}

We first evaluate end-to-end our approach's ability to generate
realistic safety-critical scenarios. \cref{sec:benchmark}
shows that compared to baselines, scenarios generated by our approach
more closely match held-out real safety-critical scenarios
both quantitatively and qualitatively.
Next, we investigate the first key aspect of our approach:
our conditional flow architecture.
We ablate our design choices and show that both conditioning and flow
are important to achieving good results (\cref{sec:ablate-model}).
The other key aspect of our approach is our data composition, and the
use of simulation data to supplement real examples.
In \cref{sec:ablate-data}, we evaluate our approach trained on various different data compositions
and show that a simple balance of simulation and real
safety-critical examples provides the best results.
Finally, we study the controllability of our approach in \cref{sec:control}.
We showcase how our model responds to the conditioning, and
how intermediate flow timesteps can be an additional lever for
controlling the specific degree of safety criticality.

\subsection{Experimental Setup}

\subsubsection{Dataset}
We conduct experiments with an in-house self-driving dataset. Our
dataset spans both highway and urban driving, consisting of approximately 20,000
traffic scenarios. Each snippet contains 20s of driving data. Approximately
10,000 snippets are from real logs, 10,000 are simulated safety critical
scenarios as described in \cref{sec:dataset}. Additionally, as described in
\cref{sec:dataset} we have curated approximately 500 real safety critical
snippets. The remaining training samples are also upsampled. We hold out 20\% of
the real safety critical data for evaluation.



\subsubsection{Metrics}

Evaluating a generative model for scenario generation is non-trivial and
requires multiple metrics measuring realism and safety-criticality. We propose our
experiments on the following suite of metrics.

\begin{itemize}

      \item \textbf{minSTTC and Near Miss Rate} To evaluate if our method learns to
            construct near-miss cases from the training distribution, we propose a
            scenario-level minimum time to collision metric (minSTTC). For each scenario
            sample, we calculate the minimal time to collision between the ego actor and the
            closest leading actor throughout the entire rollout with an upper bound of 10
            seconds. We report the median of the minSTTC because in the cases where no
            likely collision is going to happen, the minSTTC is likely to be large. We
            consider a more effective safety-critical scenario as inducing a small TTC
            without causing any collision. We also report the percentage of
            scenarios where the minSTTC is less than 3 seconds, which we consider as a near
            miss that challenges the planner.

      \item \textbf{Distribution JSD} Following common practice~\cite{igl2022symphonylearningrealisticdiverse,tan2025scenediffusercityscaletrafficsimulation},
            we compute the distributional kinematics metrics of actors. Smaller divergence indicates that
            the model captures the essence of safety-critical maneuvers. We measure against
            common kinematic metrics that characterize actor maneuver: linear and angular
            speed and acceleration. 

      \item \textbf{Collision Rate} We evaluate the average percentage of actors
            colliding in each scenario based on a small IOU threshold between the bounding
            boxes of the actors.
            Collision rate should generally be low even for safety-critical scenarios,
            since none of the ground truth data has any collisions.
            However, models can obtain low collision rate by generating nominal scenarios,
            so other metrics like minSTTC must also be considered.

      \item \textbf{Reconstruction}
            We also provide L2
            reconstruction to the ground truth trajectory as another way to measure realism
            to supplement distribution JSD.

\end{itemize}

\subsection{Generating Realistic Safety Critical Scenarios}
\label{sec:benchmark}

We compare against 3 baselines
\begin{enumerate}
      \item \emph{VAE} is the base model, trained on the same data as our approach.
      \item \emph{VAE + Curation} is the base CVAE model, trained only on safety critical scenarios (both real and sim)
      \item \emph{Strive}~\cite{rempe2022strive} is a SOTA baseline which performs optimization in latent space. We use the same
            base CVAE as our approach.
\end{enumerate}
\cref{table:benchmark} shows the results.
Compared with the baseline CVAE models, our model achieves comparable
kinematics metrics while generating more safety-critical scenarios. Our model
generates smaller minSTTC with higher near-miss rate.
Because the baseline CVAE is trained on the base mixture distribution, it does not produce
as many safety-critical scenarios, as we can see by its relatively worse minSTTC and Near Miss rate.
Our VAE + Curation baseline obtains higher near miss rate but is less realistic overall.
This is because the omitted nominal data still contains valuable learning signal,
in particular for background traffic, etc.
STRIVE also obtains a high near miss rate but suffers from realism. We
believe that this is because the prior model does not provide
strong enough regularization. Also, the adversarial optimization 
is not explicitly aware of the real world distribution of safety-critical scenarios.
On the other hand our flow approach obtains the best of both worlds
as it is able to generate a large percentage of near miss scenarios
while maintaining good performance in the other metrics.

Qualitatively (\cref{figure:demo_viz}), we can see that with an ordinary highway
scenario, our
model perturbs the maneuver of the leading actor by causing it to perform a hard
brake, leading the ego vehicle to follow as well. On the second occasion, the
model causes the actor in the neighbor lane to cut into ego vehicle's lane
aggressively, also causing a hard brake from the ego actor. One side benefit of
the flow method is that the model automatically chooses the actor to interact
with the ego vehicle as well as the maneuver to perform, so that there is no
need for explicit interaction design.

\begin{figure}[t]
    \centering
    \begin{subfigure}[h]{0.5\linewidth}
        \centering
        \includegraphics[width=\linewidth,trim={15cm 0 13cm 0},clip]{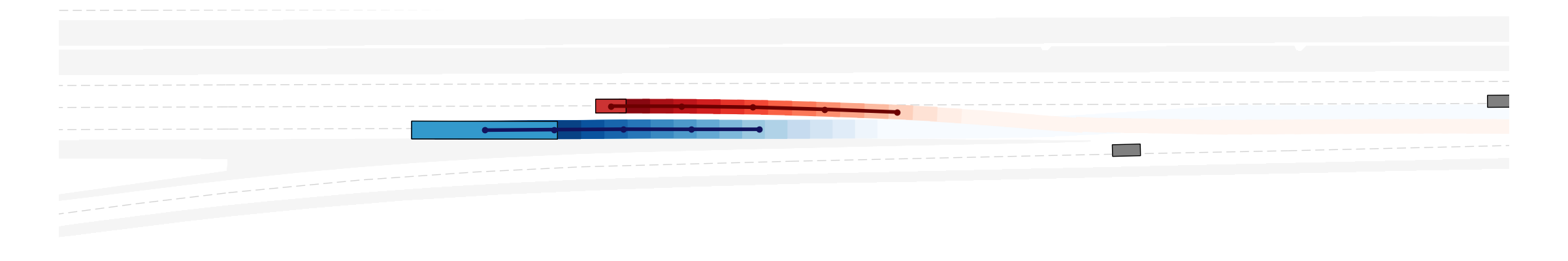}
        \includegraphics[width=\linewidth,trim={15cm 0 13cm 0},clip]{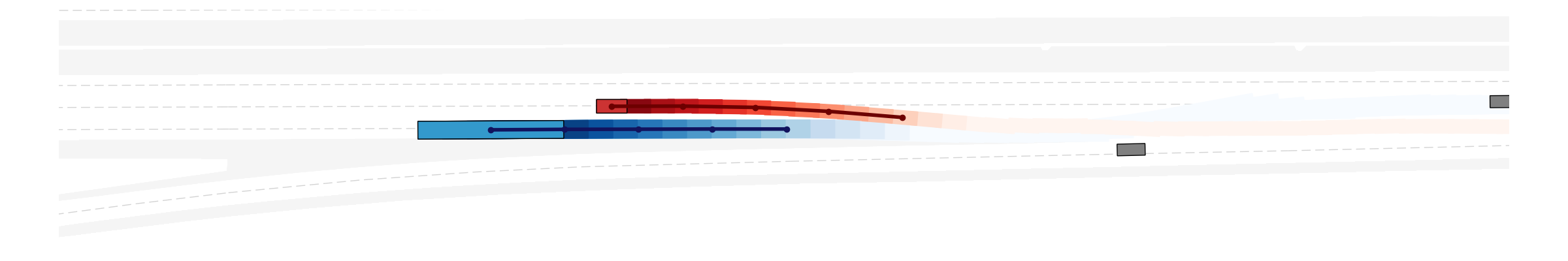}
        \includegraphics[width=\linewidth,trim={15cm 0 13cm 0},clip]{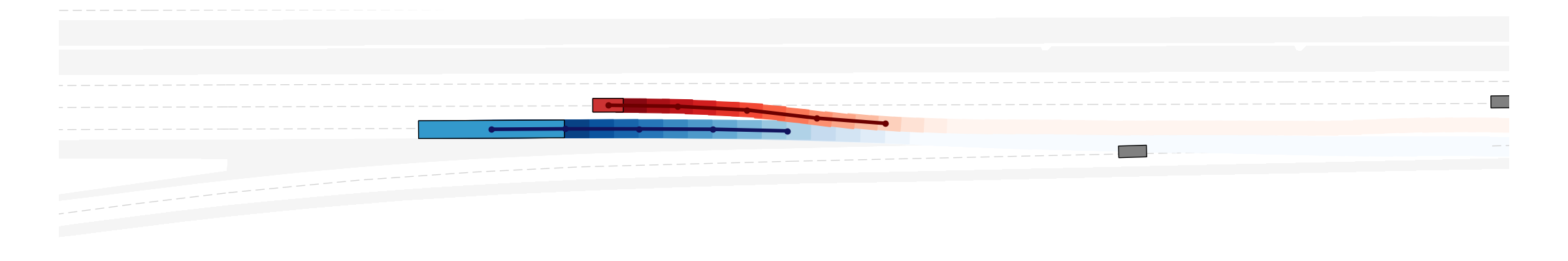}
        \caption{Cut-in scenario}
    \end{subfigure}%
    ~
    \begin{subfigure}[h]{0.5\linewidth}
        \centering
        \includegraphics[width=\linewidth,trim={15cm 0 13cm 0},clip]{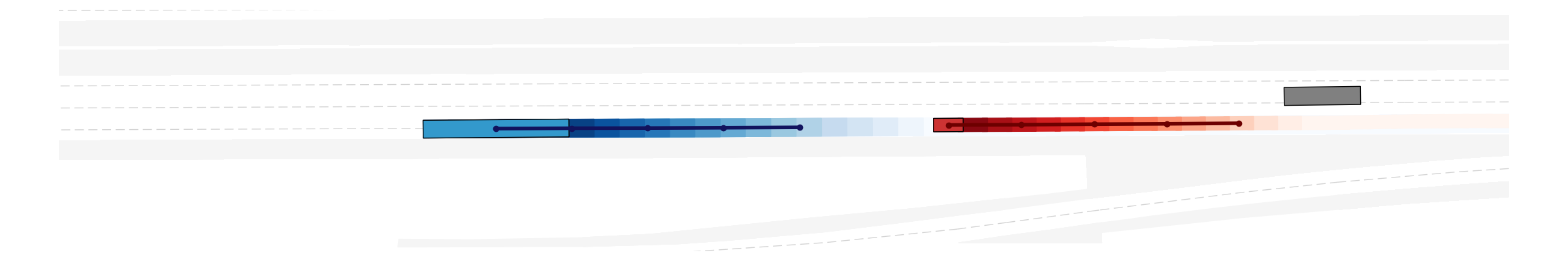}
        \includegraphics[width=\linewidth,trim={15cm 0 13cm 0},clip]{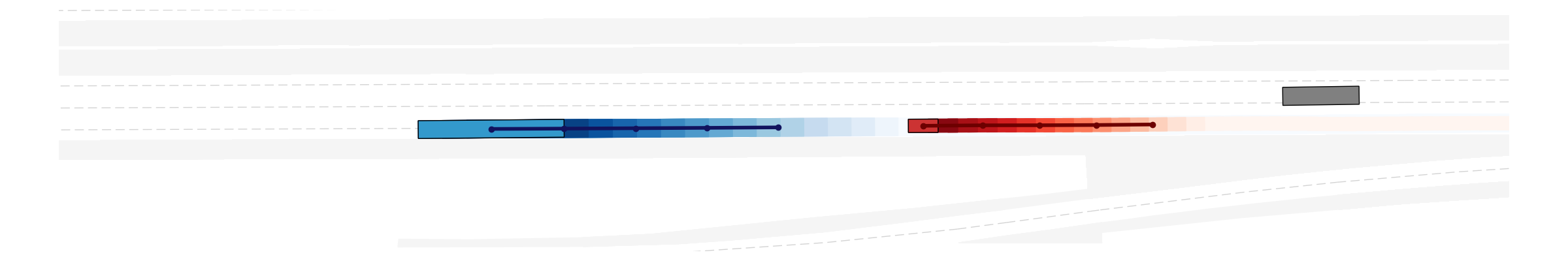}
        \includegraphics[width=\linewidth,trim={15cm 0 13cm 0},clip]{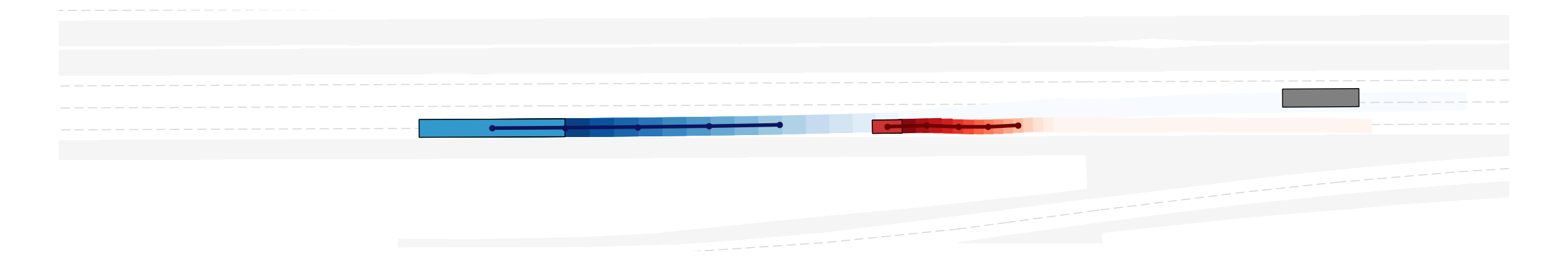}
        \caption{Hard-brake scenario}
    \end{subfigure}
    \caption{\textbf{Varying number of flow timesteps.} More flow steps 
    corresponds to higher safety criticality.}
    \label{figure:demo_cond_viz}
\end{figure}

\subsection{Conditional Flow Ablation}
\label{sec:ablate-model}
We now ablate the key architectural choices of our approach: the flow transformer,
and the additional conditioning.
The flow-only model simply removes the conditioning from the flow transformer.
For the conditioning-only model, we add a similar conditioning encoder
to the VAE prior.
\cref{table:benchmark} shows the results.
As expected, without any flow or conditioning to steer the sampling, the base model
has trouble generating safety critical rollouts. We see
that conditioning on its own is also ineffective.
Our hypothesis is that adding conditioning during the VAE training potentially
harms representation learning as it potentially provides too much of a shortcut.
On its own, flow is already effective, but adding conditioning to flow
results in the best overall model.

\subsection{Data Composition}
\label{sec:ablate-data}
We evaluate our model performance with different mixtures of
real data in \cref{table:real_mix}. For these experiments, the same
base VAE is used, but we adjust the mixed ratio of sim and
real data when training the flow model.
We see that using sim or real only is ineffective, due to
lower fidelity data, and smaller scale data respectively. Combining
them shows the best results, with a sweet spot at around 50\%.

\begin{figure}
      \includegraphics[width=\linewidth,height=4cm]{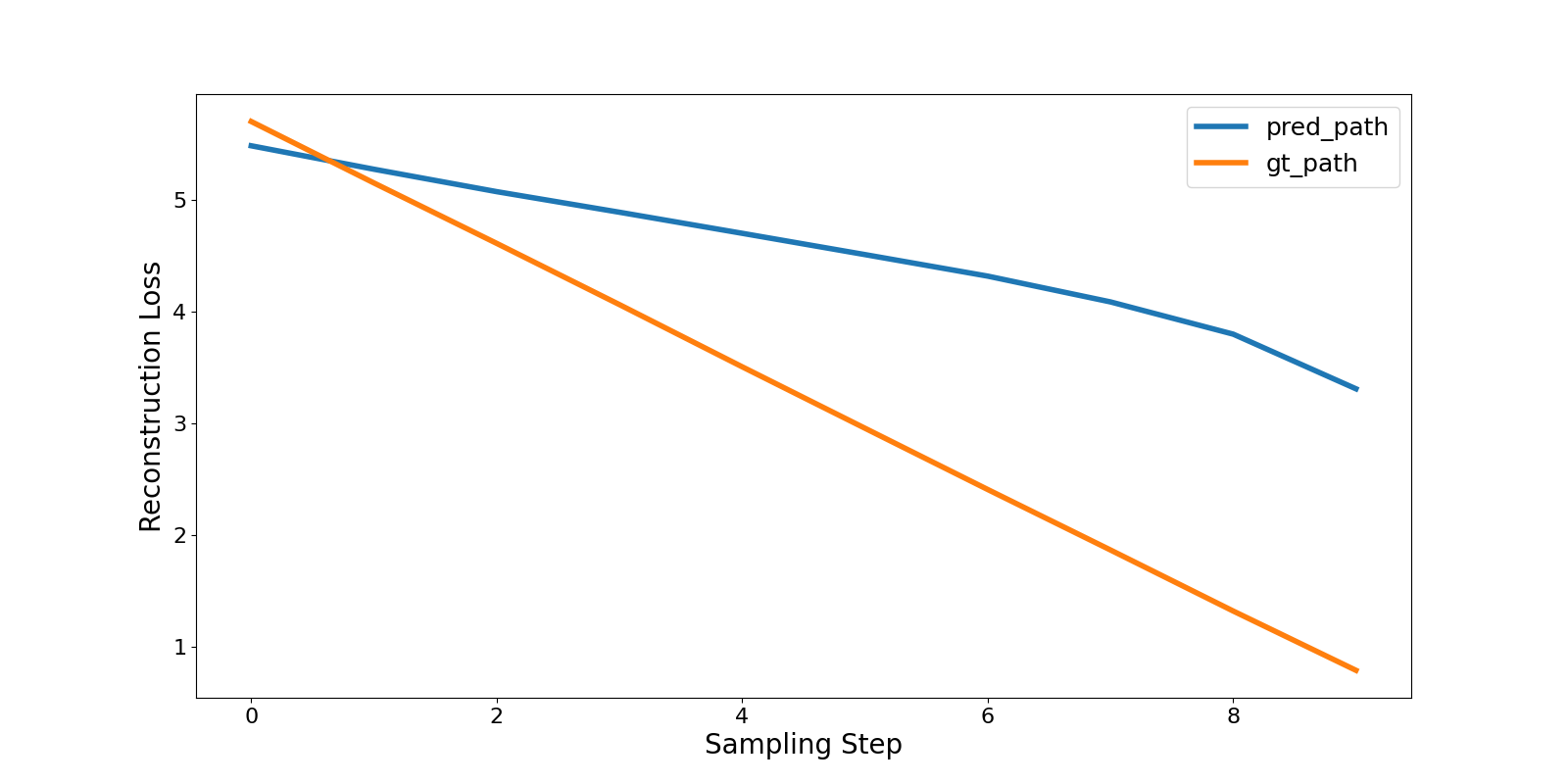}
      \caption{\textbf{Flow timesteps vs. Reconstruction Error}.
            On average, reconstruction error smoothly drops with flow timesteps.}
\end{figure}
\subsection{Controllability Studies}
\label{sec:control}
\begin{table}[t]
      \caption{
            \textbf{Varying conditioning.} We see that the model is
            controllable via the conditioning parameter.
      }
      \label{table:cond}
      \begin{center}
            \begin{tabular}{c | c c | c}
                  \hline
                                     &                      &                           &                          \\
                                     & minSTTC $\downarrow$ & Near Miss (\%) $\uparrow$ & Displ Error $\downarrow$ \\
                  \hline
                  No Cond.           & 2.190                & 0.319                     & 8.327                    \\
                  Cond. Nominal      & 3.263                & 0.281                     & 9.727                    \\
                  Cond. Intermediate & 2.293                & 0.343                     & 9.271                    \\
                  Cond. Challenging  & 1.967                & 0.500                     & 8.315                    \\
                  \hline
            \end{tabular}
      \end{center}
\end{table}

Controllability is another desired property for our model because it allows us
to generate different levels of safety-critical scenarios, which could be used
to gradually test the performance upper bound of the autonomy system. In this
part, we compare the difference of the rollouts from the same initialization but
with different maneuver labels. In \cref{table:cond}, we show the
performance of the model with different maneuver labels. We see
minSTTC decreases and near-miss rate increases as the maneuver label becomes more
challenging, while the unconditional model lies somewhere in between.
We also found that controllability could also be achieved by manipulating the
flow time steps. We sample the latents along the inference time steps and
decode them to visualize the reconstruction. \cref{figure:demo_cond_viz}
shows that the model rollout is nominal at
$t=0$, and becomes safety-critical at $t=1$,
providing more refined control of
scenario generation when used in conjunction with the maneuver labels.
\section{CONCLUSION}
In this paper, we proposed Flow VAE, a method for data-driven safety-critical
traffic scenario rollout generation. Flow VAE is a flow matching transformer
that learns to transfer the latents from nominal initialization into
safety-critical ones. We also show that using a mixture of sim and real data,
we are able to scalably generate safety-critical scenarios with a small dataset.
Our ablation studies validate our architectural and data composition design choices,
and we further show multiple methods to control our model at varying granularity.

Our real data curation and synthetic generation are proof-of-concept
and can be improved.
For instance higher fidelity vehicle models involving friction could better address sim-to-real 
and allow for more interesting scenarios involving
slipping. Scaling up data collection and synthetic generation techniques
could allow for increasingly diverse safety critical scenarios to be generated.
While we showed
controllability in the degree of safety criticality (through maneuver labels and number of flow timesteps),
controlling the maneuver itself could be an interesting future direction.



{
    \bibliographystyle{IEEEtran}
    \bibliography{refs}
}

\end{document}